\documentclass[journal,twoside,web]{IEEEtran}
\usepackage{tmi}
\usepackage{amsmath,amssymb,amsfonts}
\usepackage{algorithmic}
\usepackage{graphicx}
\usepackage{textcomp}
\usepackage{adjustbox}
\usepackage{multirow}
\usepackage{bbold}
\usepackage{siunitx}

\usepackage{colortbl}
\usepackage{hhline}
\usepackage{float}
\usepackage{diagbox}

\usepackage[normalem]{ulem}
\usepackage{color}
\usepackage{booktabs}
\usepackage{bbm}
\usepackage{makecell}
\usepackage{rotating}
\usepackage{hyperref}
\usepackage{soul}
\usepackage{mathrsfs}
\usepackage{bbding}

\usepackage{colortbl}  
\usepackage{xcolor}
\usepackage{array}

\newcommand{\change}[1]{\textcolor{black}{#1}}

\def\BibTeX{{\rm B\kern-.05em{\sc i\kern-.025em b}\kern-.08em
    T\kern-.1667em\lower.7ex\hbox{E}\kern-.125emX}}
\begin{document}
\title{Privacy-Preserving Synthetic Continual Semantic Segmentation for Robotic Surgery} 

\author{Mengya Xu, Mobarakol Islam, Long Bai and Hongliang Ren 
\thanks{The work was supported by the National Key R\&D Program of China under Grant 2018YFB1307700 (also with subprogram 2018YFB1307703) from the Ministry of Science and Technology (MOST) of China; Hong Kong Research Grants Council (RGC) Collaborative Research Fund (CRF C4026-21GF and CRF C4063-18G), and General Research Fund (GRF \#14211420); Shun Hing Institute of Advanced Engineering (BME-p1-21/8115064) at The Chinese University of Hong Kong (CUHK); and Shenzhen-Hong Kong-Macau Technology Research Programme (Type C) Grant 202108233000303 awarded to Dr. H. Ren. M. Islam was funded by EPSRC grant [EP/W00805X/1]. (M. Xu and M. Islam are co-first authors.) (Corresponding author: Hongliang Ren)}
\thanks{M. Xu and H. Ren are with the Department of Biomedical Engineering, National University of Singapore, Singapore, and NUSRI Suzhou, China. (E-mail: mengya@u.nus.edu)}
\thanks{M. Islam is with the Wellcome/EPSRC Centre for Interventional and Surgical Sciences (WEISS), University College London, UK. (E-mail: mobarakol.islam@ucl.ac.uk)}
\thanks{L. Bai and H. Ren are with the Department of Electronic Engineering, The Chinese University of Hong Kong (CUHK), Hong Kong, China. (E-mail: b.long@ieee.org)}
\thanks{H. Ren is also with Shun Hing Institute of Advanced Engineering, The Chinese University of Hong Kong (CUHK), Hong Kong, China. (E-mail: hlren@ieee.org)}
}

\maketitle
\begin{abstract}

Deep Neural Networks (DNNs) based semantic segmentation of the robotic instruments and tissues can enhance the precision of surgical activities in robot-assisted surgery. However, in biological learning, DNNs cannot learn incremental tasks over time and exhibit catastrophic forgetting, which refers to the sharp decline in performance on previously learned tasks after learning a new one. Specifically, when data scarcity is the issue, the model shows a rapid drop in performance on previously learned instruments after learning new data with new instruments. The problem becomes worse when it limits releasing the dataset of the old instruments for the old model due to privacy concerns and the unavailability of the data for the new or updated version of the instruments for the continual learning model. For this purpose, we develop a privacy-preserving synthetic continual semantic segmentation framework by blending and harmonizing (i) open-source old instruments foreground to the synthesized background without revealing real patient data in public and (ii) new instruments foreground to extensively augmented real background. To boost the balanced logit distillation from the old model to the continual learning model, we design overlapping class-aware temperature normalization (CAT) by controlling model learning utility. We also introduce multi-scale shifted-feature distillation (SD) to maintain long and short-range spatial relationships among the semantic objects where conventional short-range spatial features with limited information reduce the power of feature distillation. We demonstrate the effectiveness of our framework on the EndoVis 2017 and 2018 instrument segmentation dataset with a generalized continual learning setting. Code is available at~\url{https://github.com/XuMengyaAmy/Synthetic_CAT_SD}.

\end{abstract}


\section{Introduction}
For robot-assisted surgery, semantic segmentation of surgical instruments and tissue is a significant issue, and it is required for tracking and pose estimation of instruments in surgical scenes. Although deep learning-based methods are widely used for surgical instrument segmentation~\cite{gonzalez2020isinet}, they are unable to learn several tasks sequentially as a biological learning process. Deep neural networks (DNNs) exhibit catastrophic forgetting, which refers to the sharp decline in performance on previously learned tasks after learning a new one~\cite{mccloskey1989catastrophic}. To tackle this issue, continual learning (CL) techniques~\cite{rebuffi2017icarl,chaudhry2018riemannian} distilled the old tasks or class information while training with new tasks or classes. However, it is necessary to obtain exemplar samples from old classes to fix catastrophic forgetting. There are also exemplar-free continual learning approaches~\cite{rebuffi2017icarl,li2017lwf,douillard2020podnet,douillard2021plop} which show better learning of new tasks (plasticity) but are worse in preventing forgetting (rigidity). The scenario worsens in surgical cases, where old exemplar samples are unavailable due to licensing and privacy concerns. On the other hand, data for the new classes (in particular, new instrument classes) may also not be available in the cases of (i) new surgical instruments introduced by the vendor of the robotic system; (ii) upgrading an old instrument which yet to utilize in any surgery; and (iii) instruments usually use for rare surgery. Therefore, it is vital to design a privacy-preserving continual learning framework for robotic surgery that works within a continuously changing environment.

To handle privacy concerns, synthesizing restricted data is one of the most common attempts in the research community with medical image analysis. Mostly, it is found that the trained weights of a deep learning model can be released publicly without the original dataset due to the restriction. A generative model can synthesize the private dataset from the restricted site. Nikolenko et al.~\cite{nikolenko2021synthetic} show that synthetic medical data can be shared in the healthcare industry to promote knowledge discovery without revealing genuine patient-level data. Synthetic data is also utilized to deal with coverage gaps~\cite{liang2022advances}, privacy concerns~\cite{van2020ethical}, biases~\cite{buolamwini2018gender}, bias and imbalance in data~\cite{kortylewski2019analyzing} with the help of deep generative models.

On the other hand, the surgical scene with new instruments can be generated by blending and augmenting an instrument foreground on a surgical target background. The study~\cite{wang2022rethinking} collects open-source instrument images from publicly available surgical videos and vendors' websites and then blends them with the open-source surgical scene. In this way, it is possible to generate an instrument segmentation dataset with any desired instrument and surgical region of interest. A similar approach to instruments segmentation dataset with extensive blending techniques is presented by~\cite{garcia2021image}, where they manually capture thousands of foreground instruments and background tissue images and blend them to build the dataset for binary segmentation. These kinds of methods not only offer freedom to generate the class-balanced dataset with any desired instrument but automatically yield the instrument annotation. This is especially useful in medical applications where the annotation is expensive, time-consuming, and prone to errors.

These limitations and motivations drive us to develop a new privacy-preserving synthetic continuous semantic segmentation framework to support decision-making in robotic surgery. In this work, our framework is designed by blending and harmonizing (1) open-source old instruments foreground to the synthesized background without revealing real patient data in public, and (2) new instruments foreground to augmented real and publicly available background. Meanwhile, our methodology addresses data scarcity and the time-consuming annotation procedure. Here, open-source images refer to the instruments' foreground and tissue background images acquired from publicly available surgical videos or surgical instruments found on vendor's websites. The real test dataset serves as the ultimate testbed, and the high-fidelity simulated dataset can be utilized in robotics to facilitate faster safer learning. We propose class-aware temperature normalization (CAT) by regulating model learning capacity to improve logit distillation from the old model to the continual learning model. We also design multi-scale shifted-feature distillation (SD) to retain long and short-range spatial relationships among semantic objects to enhance the power of feature distillation. With a generalized continual learning setting, we show the efficiency of our approach on the EndoVis 2017 and 2018 instrument segmentation datasets.

We summarize our contributions in the following points:
\begin{itemize}

\item{Propose a privacy-preserving synthetic continual semantic segmentation framework that endows the model with human-like continuous learning ability and which does not need to access any real data except only one open source real background image so as not to compromise patient privacy.}

\item{Design overlapping class-aware temperature-normalization to control model learning capacity to avoid catastrophic forgetting for the non-overlapping old classes.}

\item{Introduce multi-scale shifted-feature distillation (SD) to maintain long and short-range spatial relationships among the semantic objects to enhance the power of feature distillation. }


\item{Blending and harmonizing (i) open-source old instruments foreground to the synthesized background without revealing real patient data in public and (ii) new instruments foreground to extensively augmented real background.}

\end{itemize}

The remaining paper is structured as follows: Section~\ref{sec2} introduces the related work, and Section~\ref{sec3} describes our continual learning methodology, followed by the experiment results in Section~\ref{sec4}. Finally, we draw the conclusion in Section~\ref{sec5}.

\section{Related work}
\label{sec2}
\subsection{Continual learning methods}
The design of most of the continual learning methods is inspired by the biological learning manner of plasticity, where learning new tasks and rigidity prevent catastrophic forgetting. These methods can be categorized into two major types: exemplar-based and exemplar-free. We discuss the works closely related to our proposed method in this section.

\subsubsection{Exemplar-based methods} 
Exemplar-based or rehearsal methods rely on some exemplar samples from old tasks while learning the new tasks to balance learning in both old and new tasks. The main focus is to avoid catastrophic forgetting and task-recency bias~\cite{mai2021supervised}. These methods include exemplar rehearsal, which keeps a limited number of exemplars~\cite{rebuffi2017icarl,chaudhry2018riemannian} and pseudo-rehearsal, which generates synthetic images~\cite{shin2017continual} or features~\cite{liu2020generative} as pseudo exemplars. For some applications with privacy concerns, some modern techniques~\cite{shin2017continual,wu2018incremental} do not keep old data but make up for it by building generative adversarial networks (GANs) to create images using old classes while new classes are being learned. However, most of these methods are designed to focus on classification tasks where segmentation tasks may not be compatible due to (i) the co-occurrence of pixels from multiple classes in the same image; and (ii) GANs are unable to generate corresponding segmentation masks for the exemplar classes. Replay in continual learning (RECALL)~\cite{maracani2021recall} attempt to design a continual semantic segmentation method in vision tasks by using a conditional GAN to generate the exemplar samples from the class space of past learning dataset and blend with object segmentation labels.

\subsubsection{Exemplar-free methods} 
Exemplar-free methods do not require old exemplar samples and can prevent catastrophic forgetting. Some techniques constrain the training of particular network modules to preserve old knowledge~\cite{oquab2014learning},
while others expand or modify the network architecture when adding new classes~\cite{sarwar2019incremental,xiao2014error}.
Another effective strategy to retain old knowledge is Knowledge Distillation (KD)~\cite{hinton2015distilling, bucilu2006model}. KD can maintain network stability as the network continuously learns new knowledge~\cite{rebuffi2017icarl,li2017lwf,douillard2020podnet,douillard2021plop}. Learning without forgetting (LWF)~\cite{li2017lwf} extracts old knowledge from previous models by distilling the output layer. Consecutively, LWF is extended to incremental learning techniques (ILT)~\cite{michieli2019ilt} where KD is applied on both the output layer and the intermediate feature space. Furthermore, pooled outputs distillation (POD)~\cite{douillard2020podnet} introduces pooling into distillation to improve the learning of both old and new classes. Although these methods demonstrate promising performance in continual learning, they are developed by focusing only on classification tasks that may not be compatible with semantic segmentation. In continual semantic segmentation with incremental classes, a unique issue is background shift, where pixels associated with the background may include previous classes the model has already seen and future classes that the model has not yet seen. Pseudo-label and local POD (PLOP)~\cite{douillard2021plop} adjusted the distillation loss of POD to be multi-scale and adopted pseudo-labels to deal with background shift. Most recently, generalized class incremental learning (GCIL)~\cite{mi2020generalized} allows each task to have a different number of classes, and classes that appear in different tasks can overlap. Meanwhile, the sample sizes of different classes may vary in one task. This setting is more realistic for continual learning scenarios and fits our surgical instrument semantic segmentation task.

\subsection{Image synthesis}
Several studies~\cite{shin2018medical,hamghalam2020high} have concentrated on GAN-based data synthesis, while some works~\cite{garcia2021image,wang2022rethinking} utilize image blending or image compositing to generate the synthetic images.~\cite{garcia2021image} manually gather thousands of foreground instrument images and background tissue images and use mix-blend, which refers to mixing synthetic images produced with multiple blending strategies, for example, Gaussian blending and Laplace blending, to generate the new images. Although the effort of labeling is omitted, the manual collection process of thousands of foreground/background images still makes the data generation procedure tedious and time-consuming. Compared with~\cite{garcia2021image},~\cite{wang2022rethinking} only uses one background tissue image and $2$ foreground images for each instrument, including the $2$ states of instrument clasper opening and closing respectively as data sources to eliminate the need for massive data collection. Based on these very few data sources, augmentation and blending techniques are utilized to generate composite images. DavinciGAN~\cite{lee2019davincigan} utilizes a generative adversarial network (GAN)-based approach to solving the problem of insufficient training data. Unpaired image-to-image translation technique~\cite{pfeiffer2019generating} is utilized to generate a wide range of realistic-looking synthetic images based on images from a simple laparoscopy simulation. The approach, which combines unpaired image translation with neural rendering~\cite{rivoir2021long}, is designed to transfer simulated to photorealistic surgical abdominal scenes. There are also some works using simulators to create datasets, such as a virtual surgery environment in the Unity engine, and this environment is converted photo-realistically with semantic image synthesis models~\cite{yoon2022surgical}. The kinematic data of a movement is first replicated over an animal tissue background using the dVRK, and the same kinematic data is then performed on an OLED green screen. The ground truth is produced utilizing the background subtraction approach. The collected kinematic data are then fed into a dVRK simulator to generate tool simulation images~\cite{colleoni2020synthetic}. CaRTS~\cite{ding2022carts} performs robot tool segmentation experiments on synthetic data generated using Asynchronous Multi-Body Framework (AMBF), which has been developed to facilitate fluid and seamless interaction with virtual surgical phantoms. The main distinction lies in our approach to utilizing synthetic data without relying on expensive virtual simulators, unlike the methods employed by these other works that require such assistance. Instead, we employ a cost-effective and affordable blending and harmonization technique. Compared to earlier works~\cite{garcia2021image,wang2022rethinking} focus on the simple task of binary segmentation. We use an image blending technique to synthesize images for the multiple instruments segmentation task without exorbitant data gathering and annotation costs. 

\section{Methodology}
\label{sec3}
\begin{figure}[!h]
\centering
\includegraphics[width=1\linewidth]{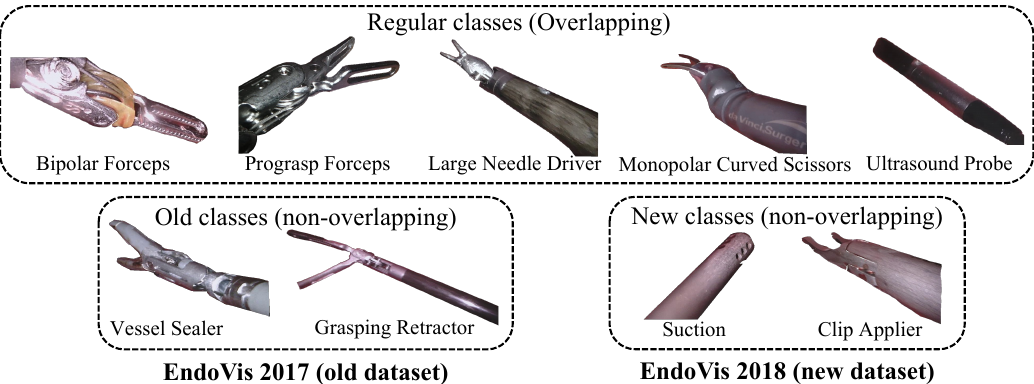}
\caption{Instruments classes in our continual learning settings. The old non-overlapping instruments in EndoVis 2017 are Vessel Sealer and Grasping Retractor, and the new non-overlapping instruments in EndoVis 2018 are Suction and Clip Applier. Other regular overlapping instruments appear in both  EndoVis 2017 and EndoVis 2018.}
\label{fig:instruments}
\end{figure}

\begin{figure*}[!hbpt]
\centering
\includegraphics[width=0.9\linewidth]{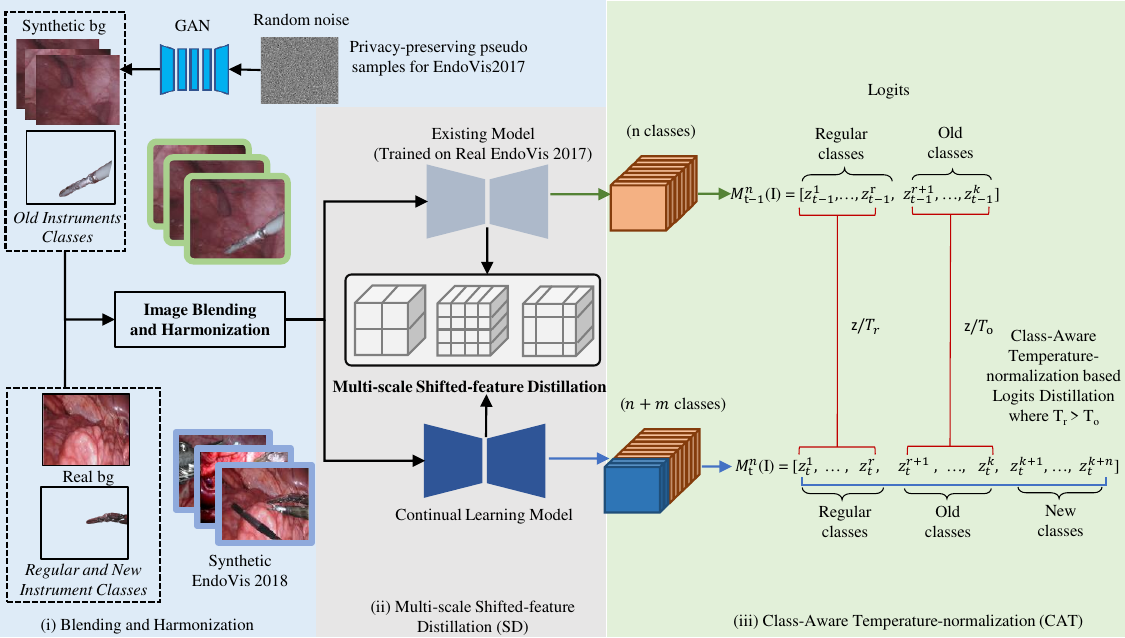}
\caption{Overview of our proposed privacy-preserving CAT-SD continual learning approach. The old model is the model weights from a hospital without sharing the training data (in our case, EndoVis 2017 dataset~\cite{allan20192017}), and it can recognize the $n$ classes. Our pseudo-rehearsal-based CAT-SD approach aims to learn a continual learning model which can deal with the m new classes from EndoVis 2018 dataset~\cite{allan20202018} and catastrophic forgetting. CAT-SD forms of modules of (i) Blending and Harmonization: to synthesize the surgical background images to blend with old non-overlapping instruments and publicly available real surgical background to blend with overlapping and new non-overlapping instruments to ensure privacy-preserving continual learning; (ii) Multi-scale Shifted-feature Distillation (SD): to enhance the feature distillation and maintain long and short-range spatial relationships among the semantic objects; (iii) Class-Aware Temperature-normalization (CAT): to tackle the imbalance learning between old and new classes based on logits distillation.
 }
\label{fig:overview}
\end{figure*}

The proposed method in this work is designed by focusing on the continual learning scenario in real-world clinical practice that requires learning incremental classes after the deployment. More specifically, training novel class categories on top of an old model or trained weights without accessing original training data. However, the novel classes are always associated with the old ones in surgical scenarios. For example, in Fig.~\ref{fig:instruments} with the instrument segmentation dataset, overall instrument classes can be divided into three groups of (i) \textit{Regular classes:} common or overlapping classes in both old and new datasets; (ii) \textit{Old classes:} non-overlapping classes in the old dataset; and (iii) \textit{New classes:} non-overlapping classes in the new dataset. We can also denote the model with old weights trained on the old dataset as \textit{old model}, and the model to train novel classes is \textit{continual learning model} as in Fig.~\ref{fig:overview}. In our case of continual learning with multiclass semantic segmentation, \textit{Old non-overlapping classes} particularly suffer from catastrophic forgetting.

The problem we address in this work is supervised multi-class continual semantic segmentation with DNNs. The goal is to train a \textit{continual learning model} at the time point $t=1$ where only the old model or other models are accessible of trained weights without the training dataset from time point $t=0$. The model can be further developed if needed for future time points $t=2,3,4,...$. In each subsequent time point, we provide a dataset $D_t$ which consists of a set of pairs $(I^t, Y^t)$, where $I^t\in\mathbb{R}^{h \times w}$ and $Y^t\in\mathbb{R}^{h \times w}$ denote an input image and corresponding ground-truth (GT) segmentation mask of size $(h, w)$. As shown in Fig.~\ref{fig:overview}, a set of $Y^t$ only contains the labels of the classes $c \in (1, ..., r, k+1,..., k+n)$ of the current time point, where (1, ..., r) indicate the regular classes and (k+1,..., k+n) indicate the new classes . The network at time point $t$ should be able to predict all classes seen so far $c \in (1, ..., n)$. $M_t$ refers to the model at time point $t$. A deep neural network at time point $t$ can be expressed as the combination of a feature extractor $f^t(.)$ and a classifier $r^t(.)$. Features can be extracted at any layer $l$ of the feature extractor ${f_l}^t, l \in \{1,...,L\}$. We refer to the output predicted segmentation mask at time point $t$ as $\hat{Y}^t = r^t \circ f^t(I)$.

\subsection{Preliminary}

\subsubsection{Continual learning with logit distillation}
A distillation loss has been applied between the \textit{old model} and the \textit{continual learning model} to prevent catastrophic forgetting in LWF for classfication~\cite{li2017lwf} and semantic segmentation in~\cite{michieli2019ilt}. This distillation loss should represent a reasonable compromise between excessive rigidity (i.e., imposing too severe limits, which prevents learning new classes) and excessive flexibility (i.e., enforcing loose constraints, which leads to catastrophic forgetting of the old classes). These techniques optimize on two losses: a cross-entropy loss ($\mathcal{L}_{CE}$), which is used for standard semantic segmentation, and a knowledge distillation (KD) loss ($\mathcal{L}_{KD}^{logits}$) applied on logits, which is used to retain the old knowledge in previous tasks. The LWF loss function, $\mathcal{L_{LWF}}=\mathcal{L}_{CE}+\alpha \mathcal{L}_{KD}^{logits}$.


At time point $t$, the loss of KD on logits is set as the masked cross-entropy loss between the output logits $z_{t-1}$ of the previous old model $M_{t-1}$ and the output logits $z_{t}$ of the current continual learning model $M_{t}$. The number of \textit{Old classes} is indicated by k. The logits distillation loss is presented below:




\begin{equation}
\label{lwf}
    \mathcal{L}_{KD}^{logits}=-\frac{1}{\left|\mathcal{D}_t\right|}  \sum_{i}^{\mathcal{D}_t}  \sum_{c}^{k} z_{t-1}^{ic} \cdot \log \left(z_{t}^{ic}\right)
\end{equation}


\subsubsection{Continual learning with feature distillation}
In addition to distilling the logits of the old and current continual learning models, the distillation schemes can also be built on intermediate features~\cite{douillard2020podnet,dhar2019learning,zhou2019m2kd}. KD loss in ILT~\cite{michieli2019ilt} is designed as the weighted addition of KD on logits and KD on intermediate features, $ \mathcal{L_{ILT}}=\mathcal{L}_{CE}+\alpha (\mathcal{L}_{KD}^{logits}+\beta \mathcal{L}_{KD}^{feature} )$. $\beta$ is a parameter used to balance the effects of KD on logits $\mathcal{L}_{KD}^{logits}$ and KD on intermediate features $\mathcal{L}_{KD}^{feature}$. $\mathcal{L}_{KD}^{feature}$ affects the intermediate features space before the decoder retains the feature learned by the previous model. As the distillation term is feature space rather than a softmax layer, the loss function is set as $L2$-norm, i.e.,

\begin{equation}
    \mathcal{L}_{KD}^{feature}=\frac{\left\|F_{t-1}-F_t\right\|_2}{\left|\mathcal{D}_t\right|}
\end{equation}

where $F_t$ represents the intermediate feature space at task $t$. ILT~\cite{michieli2019ilt} explores different loss functions, and $L2$-norm achieves the best performance. We follow this approach and adopt $L2$-norm for intermediate feature distillation in our methodology.

\subsubsection{Multi-scale pooling}

In another line of research, the feature distillation approach computes the width and height-pooled slices on features output from different layers of the old and new models at multiple scales ${s=0...S}$. In PLOP~\cite{douillard2021plop}, the embedding tensor $x \in \mathbb{R}^{c \times h \times w}$ is equally divided into $2^s$ sub-region feature embeddings at scale $s$. The POD embedding $\Theta^s(x)$ at scale $s$ is calculated as the concatenation of $2^s$ feature embeddings:
\begin{equation}
\resizebox{.85\hsize}{!}{$
    \begin{aligned}
        && \Theta^s(x) = \left[\mu\left(x_{0,0}^s\right)\ldots\, \|\mu\left(x_{0,s-1}^s\right)\ldots\, \|\mu\left(x_{s-1, s-1}^s\right)\right]\\
    \end{aligned}
$}
\end{equation}
where $x_{i,j}^s$ is a sub-region of the embedding tensor at scale $s$, $\left[ \cdot \| \cdot \right]$ denotes the concatenation operation. Each feature embedding $\mu \left( x_{i,j}^s \right)$ is the concatenation of $2$ tensors named width and height-pooled slices obtained by taking the mean of the sub-region tensor $x_{i,j}^s$ along the width axis and height axis respectively and then flattening.

When multiple scales ${s=0...S}$ are applied, the final embedding $\Theta^{Ms}(x)$ is formed by concatenating the POD embeddings $\Theta^s(x)$ of each scale $s$:
\begin{equation}
    \begin{aligned}
        && \Theta^{Ms}(x) = \left[\Theta^0(x) \| ... \| \Theta^s(x) \right], s = 0...S\\
    \end{aligned}
    \label{eq:ms}
\end{equation}

Finally, the difference of feature space at $L$ several layers from the old model and the current model is minimized by the $L2$-norm:
\begin{equation}
    \mathcal{L}_{\mathrm{LocalPOD}}=\frac{1}{L} \sum_{l=1}^L\left\|\Theta^{Ms}\left(f_l^t(I)\right)-\Theta^{Ms}\left(f_l^{t-1}(I)\right)\right\|_2
\end{equation}
where the feature embedding of the input image $I$ at the $l$th layer of the old model $f_l^{t-1}(I)$ and new model $f_l^t(I)$.

\subsubsection{Temperature normalization}
Temperature-normalization is commonly applied in knowledge distillation~\cite{hinton2015distilling}, statistical mechanics~\cite{jaynes1957information} and calibrating probabilistic models~\cite{guo2017calibration}. Temperature normalization employs a single scalar parameter $T > 0 $ for all classes. Given the logits  $z_t$, the softmax function $\sigma_{sm}$, then distillation loss from Equation~\ref{lwf} can be written as:

\begin{equation}
\label{temp_kd}
    \mathcal{L}_{KD}^{logits}=-\frac{1}{\left|\mathcal{D}_t\right|}  \sum_{i}^{\mathcal{D}_t}  \sum_{c}^{k} (z_{t-1}^{ic}/T) \cdot \log \left(z_{t}^{ic}/T\right)
\end{equation}

The value of the $T$ is determined by tuning and is always greater than $1$. A higher value of $T$ yields heavily smoothed logits.

\subsection{Privacy-preserving Synthetic Continual Semantic Segmentation} 
We develop a privacy-preserving synthetic continual semantic segmentation method in robotic surgery by designing class-aware temperature-based shifted distillation (CAT-SD) and a pseudo-rehearsal method using synthetic images to rehearse the old knowledge, namely \textbf{Synthetic CAT-SD}. As illustrated in Fig.~\ref{fig:overview}, we have (i) synthesized surgical background images to blend with old non-overlapping instruments (\textit{Synthetic pseudo-exemplars}) and open-source real surgical background images to blend with overlapping and new non-overlapping instruments (\textit{Synthetic EndoVis 2018 train dataset}) to ensure privacy-preserving continual learning; (ii) built multi-scale shifted-feature distillation (SD) to enhance the feature distillation maintain long and short-range spatial relationships among the semantic objects; (iii) tackled with imbalance learning between old and new classes, we design class-aware temperature-normalization (CAT) based logits distillation.

\subsubsection{Class-Aware Temperature-normalization (CAT)}
The key idea of our class-aware temperature-normalization (CAT) based logit distillation is to control the learning ability of the model for different classes, as shown in Fig.~\ref{fig:overview}. Previous works consider more straightforward scenarios that deviate from real scenarios, i.e., no overlap between incremental classes. However, in robotic surgery, there may be duplication of instruments used for different surgical tasks. We define these overlapping classes in old and new datasets as \textbf{\textit{Overlapping or regular classes}}. Instead of treating these different class groups equally, we propose overlapping class-aware temperature-normalization (CAT) by controlling model learning utility for different class groups. This design allows the model to devote more learning utility to the \textbf{\textit{Old classes}}. Since samples of \textbf{\textit{Regular classes}} will also appear in the new dataset, the model can devote less learning utility to avoid the recency bias. To our knowledge, existing works never employ class-wise temperature normalization to control the learning utility for the \textbf{\textit{Old classes}} to solve the problem of catastrophic forgetting. The temperature value, $T = 1$, has no effect on the learning, and $T > 1$ smoothes the logits and reduces the information learning. In the continual learning process, the learning for regular classes dominates and causes catastrophic forgetting for old classes. To tackle this, we reduce the learning capacity for regular classes over the old classes by controlling temperature normalization. The regular classes are common classes for both old and new datasets and yield higher predictions by dominating model learning. In CAT, we replace the scaler temperature value with the class-aware vector of temperatures and assign a bigger temperature value for the regular classes than the old classes. If the temperature for regular and old classes are $T_r$ and $T_o$, then our class-aware vector of temperature is:
\begin{equation}
T_{CAT} = \{T^1_o, \ .., \ T^r_o, \ T^{r+1}_r, \ ..., \ T^{k}_r\}
\end{equation}
where, we set $T_r > T_o$ and distillation loss in Equation~\ref{temp_kd} can be formulated as below.
\begin{equation}
\label{eq:cat}
    \mathcal{L}_{KD}^{logits}=-\frac{1}{\left|\mathcal{D}_t\right|}  \sum_{i}^{\mathcal{D}_t}  \sum_{c}^{k} (z_{t-1}^{ic}/T_{CAT}) \cdot \log \left(z_{t}^{ic}/T_{CAT}\right)
\end{equation}

In our implementation of CAT, the temperature value for the logits of old non-overlapping classes  $T_o=3$. For the rest of the classes, $T_r=4$. When temperature $T$ is set to 1, the CAT-based method works as the standard logit distillation. \change{Our CAT-SD improves the model's ability to learn robust features that are less vulnerable to interference.}

\subsubsection{Multi-scale Shifted-feature Distillation (SD)}
\begin{figure}[t]
\centering
\includegraphics[width=1\linewidth]{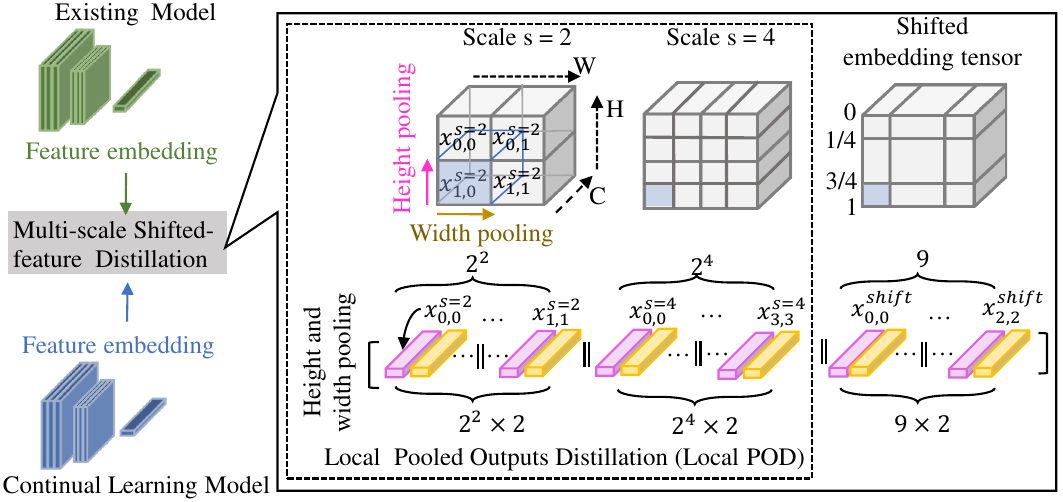}
\caption{Multi-scale Shifted-feature Distillation (SD). Two intermediate feature embeddings are obtained from the old model at task $t-1$ and the continual learning model at task $t$. The first two regular scales, $s=2$ and $s=4$ are equivalent to Local POD~\cite{douillard2021plop}. We first divide the feature embedding into $2^s$ sub-region feature embeddings equally at scale $s$. Thus $2^2$ and $2^4$ sub-regions are created separately when scales $s=2$ and $s=4$. On the basis that scales $s=4$, we group adjacent sub-regions in the interior to form irregular and unequal sub-regions, which are named the shifted embedding tensor. We then compute the width and height pooling slices for each sub-region. Eventually, all these width and height pooling slices are concatenated together. The feature distillation between the old and continual learning models is performed based on concatenated features.}
\label{fig:shifted_distillation}
\end{figure}

Motivated by the shifted window~\cite{liu2021swin}, we design the Multi-Scale Shifted Feature Distillations (SD) approach to retain long and short-range spatial linkages among semantic objects. This design overcomes the limitations of conventional short-range spatial features with a limited quantity of information restricting feature distillation's power. Firstly, as in Local POD~\cite{douillard2021plop}, we divide the spatial feature into many equal spatial feature patches based on multiple scales ${s=0...S}$, such as $s=2$ and $s=4$, represented by the first two boxes in Fig.~\ref{fig:shifted_distillation}. The same spatial feature will be equally divided into $2^2$ feature patches and $2^4$ feature patches. Then we get the regular feature tensor. Instead of combining the output feature of each layer like Local POD~\cite{douillard2021plop}, we only extract the intermediate feature space after the encoder. Because if the distillation points are set too densely, for example, every layer or neuron is used, the learning of the continual learning model may be over-regularized, which will also lead to poor distillation performance~\cite{song2022spot}. We name the regular tensor as $\Theta^{Ms}(x)$. These feature patches are called short-range spatial patches. Next, we shift the patch composition by grouping the neighbored patches to form the long-range and short-range spatial patches simultaneously, represented by the last box in Fig.~\ref{fig:shifted_distillation}. 

In our work, we use scale $s=[2,4]$ and form the shifted embedding tensor based on a scale $s$ of $4$ in our experiments. Given the embedding tensor $x \in \mathbb{R}^{c \times h \times w}$ and shifted scale $shifted \, s=4$, we introduce the shift parameter $\epsilon = [0, \frac{1}{4} , \frac{3}{4}, 1]$ to help form the shifted embedding tensor $\Theta^{Shift}_{\epsilon}$, which is described as:

\begin{equation}
        \begin{aligned}
        \Theta^{Shift}_{\epsilon}(x) = \left[\,\mu\left(x_{0,0}\right)\|\ldots\| \, \mu\left(x_{n_\epsilon-1, n_\epsilon-1}\right)\right]\\
    \end{aligned}
\end{equation}
where $n_\epsilon$ is the length of the shift parameter $\epsilon$, the irregular and unequal sub-region $x_{i, j}$ has $3$ channels and the first channel $c$ keep unchanged. $x_{i, j}=x[\,:\,, (\epsilon_{i} h:\epsilon_{i+1}h), \, (\epsilon_{j}w:\epsilon_{j+1}w) ]$,  $i \in [0, {n_\epsilon}-1]$, $j \in [{n_\epsilon}-1]$.

The shifted embedding tensor we designed breaks the fixed boundaries of these sub-regions, allowing the model to learn longer-range dependencies across the borders and short-range dependencies simultaneously. Regardless of whether the sub-region is equally or unequally divided, we compute the width and height pooling slices for each sub-region. Eventually, as shown in Figure~\ref{fig:shifted_distillation}, all these width and height pooled slices from the shifted embedding tensor $\Theta^{shift}_{\epsilon}$ and regular embedding tensors $\Theta^{Ms}(x)$ (Equation~\ref{eq:ms}) are concatenated together: 

\begin{equation}
    \Theta^{MsShift}(\Upsilon) = \left[\Theta^{Ms}(x) \| \Theta^{Shift}_{\epsilon}(x) \right]
\end{equation}

Finally, $L2$-norm is still used to minimize the difference of the feature space $\Theta^{MsShift}$ from the old model and the continual learning model, which significantly enhances the modeling and distillation power: 

\begin{equation}
\label{eq:sd}
\resizebox{.85\hsize}{!}{$
    \mathcal{L}_{SD}^{feature} = \left\|\Theta^{MsShift}\left(f_l^t(I)\right)-\Theta^{MsShift}\left(f_l^{t-1}(I)\right)\right\|_2
$}
\end{equation}


From Equations~\ref{eq:cat} and~\ref{eq:sd}, the final loss function in our CAT-SD approach is:
\begin{equation}
\label{eq:cat-sd}
    \mathcal{L}=\mathcal{L}_{CE}+ (\mathcal{L}_{CAT}^{logits}+\mathcal{L}_{SD}^{feature} )
\end{equation}

\begin{figure}[t]
\centering
\includegraphics[width=1\linewidth]{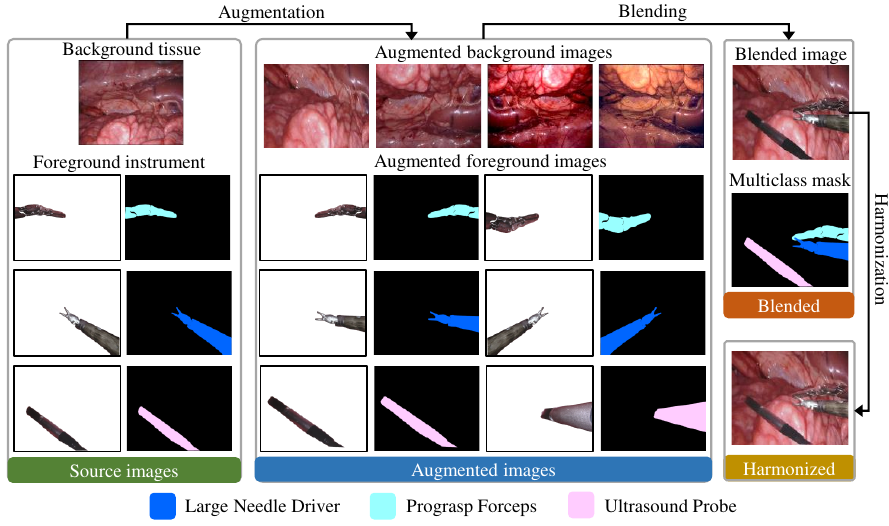}
\caption{Multi-class image synthesis process in our work consists of $4$ tasks: 1) selecting source images; 2) generating the augmented background and foreground images; 3) blending background and foreground images randomly and 4) harmonizing the blended images. One background tissue image and $2$ foreground instrument images are stored as source images for each instrument. After augmentation, the background image has $50$ variations, and the foreground image has $100$ variations. Background and foreground variations are blended randomly by limiting up to 3 tools to appear simultaneously. Eventually, the blended images are harmonized to obtain more realistic images.}
\label{fig:synthesis}
\end{figure}

\subsubsection{Blending and harmonization}
We pioneered integrating blending and harmonization techniques into the continual learning setting. Specifically, we blend and harmonize open-source old instruments foreground to the synthesized background without revealing real patient data publicly. We also blend and harmonize new foreground instrument images to extensively augment real background images to solve the problem of data scarcity and expensive and time-consuming labeling. 

\change{During the deployment phase, environmental disturbances may cause deviations in input and training data~\cite{hendrycks2019benchmarking}. Such disturbances can include variations in brightness, contrast, and other factors. Inadequate model robustness in the presence of these environmental disturbances can result in prediction errors, which can threaten the safety of surgical procedures. To address this, we adopt a data-centric approach and expose the model to a more diverse dataset to enhance its robustness.} Inspired by~\cite{garcia2021image,wang2022rethinking}, to eliminate the need for dataset collection and annotation, we introduce various augmentations to the publicly available foreground and background source images to increase dataset diversity, achieve dataset balance \change{and enhance model robustness}. For example, \change{brightness, contrast,} horizontal flip, vertical flip, rotate by $-45$ to $45$ degrees, shear by $-16$ to $16$ degrees, scale, blur, and contrast. In the blending procedure, we allow up to $3$ surgical instruments to appear in one background image simultaneously and apply to blend to the augmented foreground and background images to generate blended images and the multi-class masks, as shown in Figure~\ref{fig:synthesis}.

\change{The foreground and background appearances of images can vary considerably based on differences in image acquisition devices, brightness, and contrast, resulting in spurious features such as prominent borders of instruments. These misleading characteristics can serve as a potential source of error by providing a shortcut for the learning process, thereby confusing deep learning models during segmentation. Consequently, models trained on such data lack robustness and encounter challenges when adapting to clinical videos captured in real-life conditions. To address this problem and enhance the model's robustness,} image harmonization can be used to modify the appearance of the foreground to make it consistent with the background. A combination of encoder-decoder architectures and a pre-trained foreground-aware deep high-resolution network~\cite{sofiiuk2021foreground} is utilized to implement the image harmonization to make the blended image look realistic.

\subsubsection{Synthetic CAT-SD}
To avoid keeping exemplars and the privacy problems associated with pseudo-exemplar rehearsal, some techniques aim to generate synthetic examples for previous time points to achieve pseudo-rehearsal, which can further boost performance. However, current image generation approaches have difficulty generating complex image data realistically. Therefore, this strategy has been applied to simple datasets and is known to produce unsatisfying results on complex datasets~\cite{masana2020class}. Each instrument consists of $3$ parts: shaft, clasper, and wrist~\cite{allan20183}. The wrist and the claspers are collectively referred to as the head. However, the shafts of many robotic surgical instruments are very similar. Only the heads of these instruments are distinct. Due to the limited surgical field of view and the manipulation of instruments on organs/tissues, the head of instruments retains a tilted posture, making it challenging to capture the head features. Additionally, the head part utilized to distinguish different surgical tools are frequently occluded by organs/tissues due to their surgical duties. For the above reasons, generating various and realistic surgical instruments is a very challenging task. 

\begin{figure}[!h]
\centering
\includegraphics[width=0.9\linewidth]{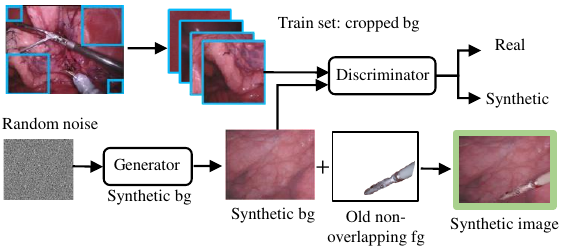}
\caption{Privacy-preserving pseudo-exemplar. When presented with the cropped background instances from the real dataset, the discriminator should recognize genuine ones. Meanwhile, the generator generates synthetic background images that it sends to the discriminator. In pseudo rehearsal, the foreground instruments are blended with the background tissue generated by the GAN model.}
\label{fig:privacy_preserving}
\end{figure}

Moreover, the tissue, as the background in the robotic surgery, is considered the patient's private data. To address the privacy restrictions imposed, we utilize StyleGAN-XL~\cite{sauer2022stylegan}, consisting of the generator and the discriminator neural networks, to generate the tissue image as background. The generator neural network generates the synthetic background images, while the discriminator neural network evaluates them for authenticity. The objective of the generator is to create lifelike background tissue images without getting caught by the discriminator. For privacy preservation, we only need to obtain the generator neural network weights of the StyleGAN-XL for the subsequent time points. Inspired by~\cite{garcia2021image, wang2022rethinking}, we utilize the cropped instruments as foreground, which does not contain any privacy of the patient. Then we blend these instruments with the synthetic tissue background output from the generator of the StyleGAN-XL~\cite{sauer2022stylegan} to create the images without revealing the patient privacy (see Figure~\ref{fig:privacy_preserving}).

\section{Experiments}
\label{sec4}
\subsection{Dataset}
\textbf{The EndoVis 2017 Dataset} is from the Robotic Instrument Segmentation Sub-Challenge\footnote{\url{https://endovissub2017-roboticinstrumentsegmentation.grand-challenge.org/}} of the Endoscopic Vision Challenge 2017~\cite{allan20192017}. The official train set includes $8 \times 225$-frame robotic surgical videos, and the official test set contains the $8 \times 75$-frame videos and $2 \times 300$-frame videos. We split the entire dataset based on the inter-video setting, e.g., video sequence number) to ensure the model has no prior information regarding the instrument in the test set. Specifically, the surgical sequence $[2,5,9,10]$ is chosen as the test set, and the rest of the sequences are used as the training set. Because the surgical sequences $[1,2]$ in the official test set do not provide the segmentation label for ``Ultrasound Probe", we ignore such $2 \times 75$ frame videos.

\begin{figure}[!hbpt]
\centering
\includegraphics[width=0.9\linewidth]{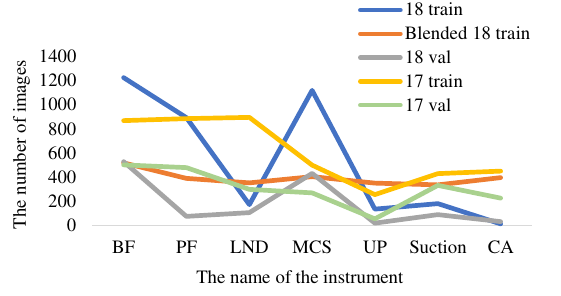}
\caption{Instruments class distribution of real and our blended EndoVis2018 train dataset.}
\label{fig:statistics2}
\end{figure}

\textbf{The EndoVis 2018 Dataset} is from the Robotic Scene Segmentation Sub-Challenge\footnote{\url{https://endovissub2018-roboticscenesegmentation.grand-challenge.org/home/}} of the Endoscopic Vision Challenge 2018~\cite{allan20202018}. The released dataset includes $15$ robotic surgical videos. The surgical sequence $[2,5,9,15]$ is used as a test set, and the rest of the sequences are selected as the training set by following~\cite{gonzalez2020isinet}. Figure~\ref{fig:statistics2} demonstrates the statistics of our training and validation set of EndoVis 2017 and EndoVis 2018 and our synthetic EndoVis 2018 training set. The flat line (\textit{Blended 18 train}) in Figure~\ref{fig:statistics2} indicates that all classes in the dataset are balanced. On the contrary, if the polyline fluctuates significantly, the dataset classes are seriously unbalanced. The medical field suffers from a severe shortage of data. The four polylines representing the training and validation set of EndiVis 2017 and 2018 are very choppy. The scarcity of data motivates us to design the synthetic CAT-SD for the continual learning problem under the unbalanced data.



As shown in Figure~\ref{fig:instruments}, the old non-overlapping instruments or old classes that appear in EndoVis 2017 are Vessel Sealer and Grasping Retractor. The new non-overlapping instruments or new classes that appear in EndoVis 2018 are Suction and Clip Applier. Other overlapping instruments appear in both  EndoVis 2017 and EndoVis 2018. In the continual learning perspective, \textbf{\textit{Overlapping or regular classes}} are bipolar forceps, prograsp forceps, large needle driver, monopolar curved scissors, and ultrasound probe. \textbf{\textit{Old classes}} are vessel sealer and grasping retractor. \textbf{\textit{New classes}} are suction and clip applier.

\subsection{Experiments setting} 

In our continual learning setting, we assume that we can only have access to the old model weights (the time point $t=0$) trained on EndoVis 2017 dataset, and we cannot access the EndoVis 2017 train set due to privacy. The continual learning model (the time point $t=1$) only has access to the EndoVis 2018 dataset, which includes new instruments. The test set is the combination of the test set of the EndoVis 2017 and 2018 datasets.

\subsection{Implementation details}
We use the PyTorch framework to implement our models and train all models with an NVIDIA RTX3090 GPU. For efficient computing, images and corresponding masks are resized into $224 \times 224$. The models are trained with a learning rate of $0.01$ and $0.001$ for the time points $t=0$ and $t=1$, respectively. The best model is selected based on the best mean Intersection over Union (IoU) performance on the validation set for each time point. The best model from the previous time point is utilized as a starting point for the subsequent time points. 

The old model weights ($t = 0$) and the continual learning model ($t = 1$) share the  same architecture: DeepLabv3+~\cite{chen2017rethinking} with a ResNet-101~\cite{he2016deep} backbone. The classification layers of the old model and the continual learning model have different numbers of output neurons based on new classes. During continual learning training, the old model with trained weights works as a teacher model to produce logits for the distillation loss. 
Both old and continual learning models are initialized with the weights of the old model, and only the continual learning model is optimized with standard cross-entropy loss and CAT-SD distillation losses. We adopt the official implementation of reference continual learning methods such as LwF, ILT, POD, and PLOP by the following code~\footnote{\url{https://github.com/arthurdouillard/CVPR2021_PLOP/}}.

\begin{table*}
\centering
\caption{Performance of continual semantic segmentation on the overall and individual test set of the EndoVis 2017 and 2018. We report the approaches without continual learning (WO CL) and those with continual learning (W CL). WO CL approaches include the old model on the time point $t=0$, the naive fine-tuning (FT) baseline without forgetting mitigation, and other methods of W CL after training on the time point $t=1$.}
\scalebox{0.9}{
\begin{tabular}{c|c|c|c|c|c|c|c|c|c} 
\hline
\multirow{3}{*}{}      & \multirow{3}{*}{Approach} & \multicolumn{6}{c|}{EndoVis
  2017 + EndoVis 2018}                                                                                                                                                                                      & EndoVis 2017~  & EndoVis 2018~   \\ 
\cline{3-10}
                       &                           & Regular classes & \multicolumn{2}{c|}{Old Classes}                                                                               & \multicolumn{2}{c|}{New Classes}                                & All classes    & All classes    & All classes     \\ 
\cline{3-10}
                       &                           & mIoU            & \begin{tabular}[c]{@{}c@{}}Vessel\\ Sealer\end{tabular} & \begin{tabular}[c]{@{}c@{}}Grasping\\ Retractor\end{tabular} & Suction        & \begin{tabular}[c]{@{}c@{}}Clip\\ Applier\end{tabular} & mIoU       & mIoU       & mIoU        \\ 
\hline
\multirow{2}{*}{W/O CL} & Old model (t = 0)                 & 47.57               & 17.88                                                   & 4.82                                                         & 0.00           & 0.00                                                   & 30.81          & 43.04          & 28.65           \\
                       & FT                        & 51.03               & 0.72                                                    & 2.59                                                         & 5.51           & 0.00                                                   & 31.50          & 36.13          & 34.16           \\ 
                      & Split model                        & 47.28               & 0.00                                                    & 0.00                                                         & 3.76           & 0.00                                                   & 28.50          & 29.31         &  30.01           \\ 
                        \hline
\multirow{9}{*}{W CL}  & PI~\cite{zenke2017si}                        & 50.92               & 0.73                                                    & 2.54                                                         & 5.07           & 0.00                                                   & 31.38          & 36.11          & 33.98           \\
                       & R-Walk~\cite{chaudhry2018riemannian}                    & \textbf{52.97}      & 9.41                                                    & \textbf{6.68}                                                & 0.00           & 0.00                                                   & 33.39          & 41.16          & 32.37           \\
                       & LWF~\cite{li2017lwf}                       & 50.03               & 20.46                                                   & 4.55                                                         & 0.00           & 0.00                                                   & 32.52          & \textbf{45.04} & 29.88           \\
                       & ILT~\cite{michieli2019ilt}                       & 50.18               & 17.70                                                   & 4.04                                                         & 0.00           & 0.00                                                   & 32.28          & \textbf{43.97}          & 30.25           \\
                       & POD~\cite{douillard2020podnet}                   & 50.62               & 8.32                                                    & 5.32                                                         & 0.00           & 0.00                                                   & 31.73          &   37.41	
             &   31.33              \\
                       & Local POD~\cite{douillard2020podnet}                & 49.94               & 7.95                                                    & 5.74                                                         & 0.00           & 0.00                                                   & 31.33          &    38.37	
            &  32.02               \\
                       & LWF-MC~\cite{rebuffi2017icarl}                   & 49.70               & 21.04                                                   & 4.32                                                         & 0.00           & 0.00                                                   & 32.36          & 43.22          & 30.79           \\ 
\cline{2-10}
                       & Our Synthetic CAT-SD      & 49.96               & \textbf{23.36}                                          & 5.05                                                         & \textbf{16.48} & \textbf{4.67}                                          & \textbf{34.93} & 43.66          & \textbf{34.56}  \\
\hline
\end{tabular}
}

\label{table:main}
\end{table*}

\subsection{Experimental results}

We conduct extensive experiments and evaluations to validate our proposed approach and investigate the effectiveness of privacy-preserving continual learning in robotic surgery. We use mean intersection over union (mIoU) to measure the segmentation predictions for all the experiments across the different class groups of regular, old, new, and overall classes. In the task of surgical background synthesis, we adopt the commonly used Fréchet inception distance (FID), kernel inception distance (KID), and precision-recall from~\cite{kamil2019literature} to evaluate the quality of image generation. Our generated surgical background images (see Fig.~\ref{fig:privacy_preserving}) achieve KID of $0.03$, FID of $91.95$, precision of $0.34$, and recall of $0.04$, which indicates that our generated images are very similar to real surgical background images in color and texture. 

\begin{figure}[!h]
\centering
\includegraphics[width=1\linewidth]{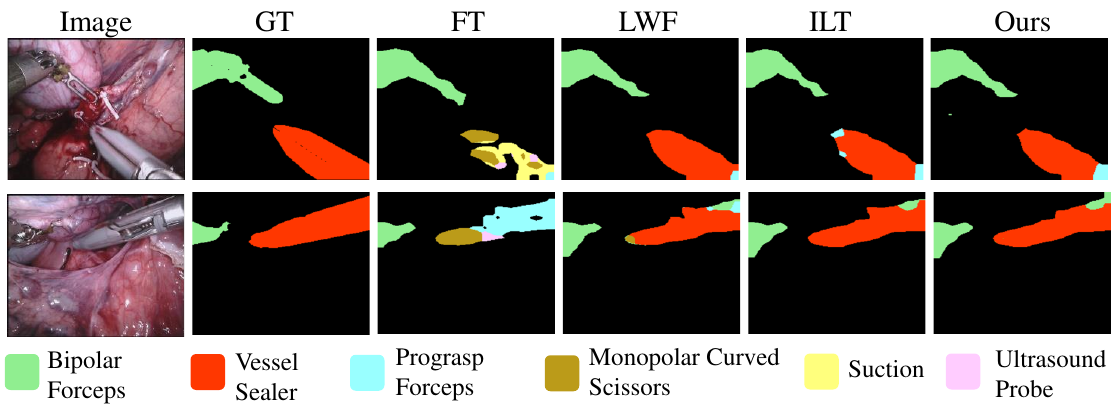}
\caption{Visualization of predicted segmentation with our method over different baselines. Different color areas represent different class. The red, light green and light blue indicate the instruments of vessel sealer, bipolar forceps and prograsp forceps.}
\label{fig:predicted_mask_v2}
\end{figure}

Our experimental results and validation studies are presented in the TABLES~\ref{table:main},~\ref{tab:synthetic},~\ref{tab:ab_2} and Fig.~\ref{fig:predicted_mask_v2}. The performance of our method is compared with the current state-of-the-art CL methods in the class categories of regular, old, and new instrument classes (definition in the Section~\ref{sec3} and Fig.~\ref{fig:instruments}), including overall performance in the test sets of EndVis 2017 and 2018 in the TABLE~\ref{table:main}. The prediction capacity on old classes (\textit{vessel sealer} and \textit{grasping retractor}) carries the evidence of catastrophic forgetting. The naive fine-tuning (FT) without the support of any CL technique presents a clear catastrophic forgetting compared to the old model. Here, the split mode refers to training the model on the EndoVis 2018 data set and not loading the checkpoint pre-trained on the EndoVis 2017 data set; that is, no fine-tuning technology is used. Specifically, after fine-tuning (FT), i.e., without any forgetting mitigation strategy, the model's prediction performance on the non-overlapping old class severely degrades, especially for the vessel sealer, which decay almost from 17.88 to 0. This result has demonstrated that the model suffers from catastrophic forgetting. PT, R-Walk, LWF, ILT, POD, Local POD, and LWF-MC approaches are proposed to alleviate this catastrophic forgetting. Compared with the results of FT, the model's prediction performance on these non-overlapping old classes is improved more, which proves that the forgetting has been alleviated more. On the other hand, most CL techniques try to preserve the old knowledge for those classes where our synthetic CAT-SD demonstrates significant improvement over other techniques. Our method not only preserves the knowledge of the old classes but also enhances them. For example, the \textit{vessel sealer} class obtains around 2-5\% improved prediction over the best CL reference model, LWF-MC, and without the CL model (old model). 

On the other hand, the performance of the new classes (\textit{suction} and \textit{clip applier}) is almost zero for the methods of w/o CL and baselines with CL. Although the methods with CL expect to learn new classes by preserving performance for the old classes, the frequency of the new classes is extremely low (as shown in Fig.~\ref{fig:statistics2}), which is very realistic in many medical datasets. In this scenario, our synthetic CAT-SD successfully learns new classes by avoiding catastrophic forgetting of the old classes. There are also exciting cases in the performance of the overall old (EndoVis 2017) and new datasets (EndoVis 2018). Our method obtains a balance performance in both datasets where most of the baselines perform poorly either in the old or new dataset. We also present a qualitative comparison in Fig.~\ref{fig:predicted_mask_v2}. Our method's segmentation predictions are less false positive and true negative and aligned with the ground truth (GT).

To ensure a fair comparison, we also introduce synthetic pseudo-exemplar for the best-performing baseline of ILT as shown in the TABLE~\ref{tab:synthetic}. Although ILT with synthetic data produces competitive performance in learning new classes (specifically with the \textit{clip applier} class), ours still shows 1-2\% higher prediction mIoU for the old classes to mitigate the catastrophic forgetting better. The performance improvement of the ILT with synthetic over vanilla ILT shows that including synthetic data improves the model prediction without costing additional data collection and annotation.

\begin{table}[!h]
\caption{Performance of continual semantic segmentation with our synthetic EndoVis2018 train dataset.}
\label{tab:my-table}
\scalebox{0.85}{
\begin{tabular}{c|c|cc|cc|c}
\hline
\multirow{2}{*}{Approach} & Regular & \multicolumn{2}{c|}{Old Classes}                                                                                                              & \multicolumn{2}{c|}{New Classes}                                           & {All classes}       \\ \cline{2-7} 
                          & mIoU    & \multicolumn{1}{c|}{\begin{tabular}[c]{@{}c@{}}Vessel\\  Sealer\end{tabular}} & \begin{tabular}[c]{@{}c@{}}Grasping\\  Retractor\end{tabular} & \multicolumn{1}{c|}{Suction} & \begin{tabular}[c]{@{}c@{}}Clip\\  Applier\end{tabular} & mIoU  \\ \hline
FT                        & 46.6    & \multicolumn{1}{c|}{0.00}                                                        & 0                                                             & \multicolumn{1}{c|}{20.4}    & 7.86                                                    & 30.79 \\ \hline
PI                        & 46.2    & \multicolumn{1}{c|}{0.00}                                                        & 0.00                                                             & \multicolumn{1}{c|}{18.33}   & 8.10                                                     & 30.36 \\
LWF                       & 47.31   & \multicolumn{1}{c|}{15.16}                                                    & 1.77                                                          & \multicolumn{1}{c|}{10.25}   & 4.30                                                     & 31.53 \\
ILT                       & 48.83   & \multicolumn{1}{c|}{19.09}                                                    & 5.21                                                          & \multicolumn{1}{c|}{10.80}    & 9.31                                                    & 33.74 \\ \hline
ILT+Synthetic             & 48.91   & \multicolumn{1}{c|}{21.61}                                                    & 4.40                                                           & \multicolumn{1}{c|}{15.14}   & 5.35                                                    & 33.99 \\
Ours          & 49.96   & \multicolumn{1}{c|}{23.36}                                                    & 5.05                                                          & \multicolumn{1}{c|}{16.48}   & 4.67                                                    & \textbf{34.93} \\ \hline
\end{tabular}
}
\label{tab:synthetic}
\end{table}

\subsection{Ablation Study}
We present the ablation study conducted to investigate the impact of hyper-parameters of our CAT-SD approach, specifically the temperature parameters $To$ and $Tr$ of CAT (see Table~\ref{tab:ablation_CAT}) and the scaling parameters $s$ and $shifted \, s$ of SD (see Table~\ref{tab:ablation_SD}). \change{To disable the CAT method, we set the temperature parameters $T_o$ and $T_r$ to 1. We set the scale parameter $s$ to 1 to disable the SD method.} In the CAT scheme, we investigate various strategies for $T_o$ and $T_r$ (see TABLE~\ref{tab:ablation_CAT}). \change{Disabling the CAT method and setting both parameters $T_o$ and $T_r$ to 1 results in the lowest performance. However, enabling the CAT method enhances the learning performance of the class.} Through experimentation, we find that setting $T_o=3$ and $T_r=4$ yields the best performance for the ``Vessel Sealer" and additionally exhibits satisfactory performance for the ``Grasping Retractor". Using a smaller $T_o$, compared to $T_r$, encourages the model to distill more information from the old classes. In the SD approach (see Table~\ref{tab:ablation_SD}), \change{when the SD method is disabled by setting the scale parameter $s$ set to 1, performance decreases to its minimum. Conversely, enabling the CAT method enhances class learning performance.} The parameter set $s=2, 4$ and $shifted \, s =4$ demonstrates superior performance in both regular classes and old classes.

We generated two sets of tissue background images using StyleGAN-XL~\cite{sauer2022stylegan} and DDPM (Denoising diffusion probabilistic model)~\cite{ho2020denoising}. We then blended these background images with the foreground old instrument images, harmonized the blended images, and consequently generated pseudo-sample sets. Experimental results are presented in TABLE~\ref{tab:gan_vs_diffusion} when our Synthetic CAT-SD method is applied with pseudo-samples obtained from StyleGAN-XL~\cite{sauer2022stylegan} and DDPM~\cite{ho2020denoising}, respectively. The ability to predict on old classes, such as the "vessel sealer" and "grasping retractor" provides substantial evidence of catastrophic forgetting. Specifically, the better the prediction results on the old class indicate that this method is more effective in overcoming and mitigating catastrophic forgetting. In Table~\ref{tab:gan_vs_diffusion}, DDPM-based Synthetic CAT-SD is 4.93\% higher than StyleGAN-XL-based Synthetic CAT-SD. However, StyleGAN-XL-based Synthetic CAT-SD is 3\% higher than DDPM-based approach. Judging from the performance averages on the old class, the results of the two methods are close. This shows that using different methods to synthesize background images has little impact on the results. This is consistent with the conclusion that the quality of the background images does not seriously affect the segmentation performance of the foreground surgical instruments, as demonstrated in the work by Wang et al.~\cite{wang2022rethinking}.

\begin{table}[htbp]
  \centering
  \caption{Ablation study of temperature $T_o$ and $T_r$ in the CAT approach.}
  \scalebox{0.9}{
    \begin{tabular}{cccccc}
    \toprule
    \multicolumn{2}{c}{CAT Temperature} & Regular classes & \multicolumn{2}{c}{Old classes} & All classes \\
    \midrule
    To    & Tr    & mIoU  & \makecell{Vessel\\Sealer} & \makecell{Grasping\\Retractor} & mIoU \\
    \midrule
     1.0   & 1.0   & 50.10  & 16.54 & 3.64 & 31.26 \\ 
    2.0   & 4.0   & 52.83 & 20.49 & 4.65 & \textbf{34.21} \\
    3.0   & 4.0   & 52.16 & \textbf{22.40} & 4.67 & 34.19 \\
    4.0   & 3.0   & 52.78 & 20.55 & 4.13 & 33.95 \\
    4.0   & 5.0   & 52.70 & 17.14 & 4.96 & 33.83 \\
    4.0   & 2.0   & 50.90 & 21.08 & 3.45 & 32.99 \\
    2.0   & 5.0   & 50.18 & 17.72 & 4.02 & 32.28 \\
    3.0   & 5.0   & 52.73 & 16.97 & \textbf{4.99} & 33.83 \\
    3.5   & 5.0   & 52.74 & 17.00 & 4.97 & 33.84 \\
    3.0   & 4.5   & 52.84 & 18.92 & 4.82 & 34.08 \\
    3.5   & 4.5   & \textbf{52.84} & 18.84 & 4.86 & 34.07 \\
    \bottomrule
    \end{tabular}%
    }
  \label{tab:ablation_CAT}%
\end{table}%

\begin{table*}[htbp]
  \centering
  \caption{The results obtained when employing different methods StyleGAN-XL~\cite{sauer2022stylegan} and  DDPM~\cite{ho2020denoising} for synthesizing tissue background images}
  \scalebox{0.9}{
    \begin{tabular}{ccccccccc}
    \toprule
    \multicolumn{1}{c|}{\multirow{3}[6]{*}{Background source}} & \multicolumn{6}{c|}{EndoVis 2017 + EndoVis 2018} & \multicolumn{1}{c|}{EndoVis 2017 } & EndoVis 2018  \\
\cmidrule{2-7}    \multicolumn{1}{c|}{} & \multicolumn{1}{c|}{Regular classes} & \multicolumn{2}{c|}{Old classes} & \multicolumn{2}{c|}{New classes} & \multicolumn{1}{c|}{All classes} & \multicolumn{1}{c|}{All classes} & All classes \\
\cmidrule{2-9}    \multicolumn{1}{c|}{} & \multicolumn{1}{c|}{Mean IoU} & \multicolumn{1}{p{6em}}{Vessel\newline{}Sealer} & \multicolumn{1}{p{6em}|}{Grasping\newline{}Retractor} & Suction & \multicolumn{1}{p{6em}|}{Clip\newline{}Applier} & \multicolumn{1}{c|}{Mean IoU} & \multicolumn{1}{c|}{Mean IoU} & Mean IoU \\
    \midrule
    StyleGAN-XL~\cite{sauer2022stylegan}  & \textbf{49.96} & \textbf{23.36} & 5.05  & 16.48 & 4.67  & \textbf{34.93} & \textbf{43.66} & \textbf{34.56} \\
    DDOM~\cite{ho2020denoising} & 45.65 & 18.43 & \textbf{8.08} & \textbf{20.82} & \textbf{8.08} & 32.93 & 40.06 & 31.55 \\
    \bottomrule
    \end{tabular}%
    }
  \label{tab:gan_vs_diffusion}%
\end{table*}%

\begin{table}[htbp]
  \centering
  \caption{Ablation study of scale $s$ and shifted scale $shifted$ $s$ in the SD approach.}
  \scalebox{0.9}{
    \begin{tabular}{cccccc}
    \toprule
    \multicolumn{2}{c}{SD Scale} & Regular classes & \multicolumn{2}{c}{Old classes} & All classes \\
    \midrule
    $s$     & $shifted \, s$ & mIoU  & \makecell{Vessel\\Sealer} & \makecell{Grasping\\Retractor} & mIoU \\
    \midrule
    1 & \XSolidBrush &  48.51 & 17.45 & 2.26  & 31.03 \\
    1, 2  & 4     & 49.64 & 18.81 & 3.76 & 32.04 \\
    1, 2  & 2, 4  & 49.60 & 18.83 & \textbf{3.77} & 32.02 \\
    1, 2, 4 & 4     & 49.80 & 19.12 & 3.35 & 32.13 \\
    1, 2, 4 & 2, 4  & 49.77 & 19.10 & 3.45 & 32.12 \\
    2, 4, 8 & 4     & 49.80 & 19.14 & 2.74 & 32.07 \\
    2, 4, 8 & 2, 4  & 49.90 & 19.25 & 3.05 & 32.17 \\
    1, 2, 4, 8 & 4     & 49.82 & 19.13 & 2.89 & 32.09 \\
    1, 2, 4, 8 & 2, 4  & 49.91 & 19.29 & 3.11 & 32.19 \\
    2, 4  & 4     & \textbf{50.18} & \textbf{20.94} & 2.58 & \textbf{32.46} \\
    \bottomrule
    \end{tabular}%
    }
  \label{tab:ablation_SD}%
\end{table}%

\begin{table}
\centering
\caption{Ablation study of our Synthetic CAT-SD. En18 and Exe mean EndoVis 2018 train dataset and pseudo-exemplars. VS, GR, and CA mean vessel sealer, grasping retractor, and clip applier.}
\scalebox{0.87}{
\centering
\begin{tabular}{c|c|c|c|c|c|c|c|c} 
\hline
\multirow{2}{*}{SD}         & \multirow{2}{*}{CAT}        & \multicolumn{2}{c|}{Synthetic}                            & Regular & \multicolumn{2}{c|}{Old Classes} & \multicolumn{2}{c}{New Classes}  \\ 
\cline{3-9}
                            &                             & En18                      & Exe                   & mIoU    & VS    & GR                       & Suction & CA                     \\

\hline
\XSolidBrush   & \XSolidBrush   & \XSolidBrush & \XSolidBrush & 50.18   & 17.70 & 4.04                     & 0.00    & 0.00                   \\ 

\hline
\Checkmark   & \Checkmark   & \XSolidBrush & \XSolidBrush & \textbf{52.92}   & 20.62 & 4.38                     & 0.00    & 0.00                   \\ 

\hline
\XSolidBrush & \XSolidBrush & \Checkmark   & \XSolidBrush & 48.83   & 19.09 & 5.21                     & 10.80   & \textbf{9.31}                   \\ 
\hline
\Checkmark   & \XSolidBrush & \Checkmark   & \XSolidBrush & 49.36   & 22.13 & \textbf{5.48}                     & 14.46   & 6.78                   \\ 
\hline
\Checkmark   & \Checkmark   & \Checkmark   & \XSolidBrush & 49.21   & 22.78 & 5.42                     & 15.51   & 5.68                   \\ 
\hline
\Checkmark   & \Checkmark   & \Checkmark   & \Checkmark   & 49.96   & \textbf{23.36} & 5.05                     & \textbf{16.48}   & 4.67                   \\
\hline
\end{tabular}
}
\label{tab:ab_2}
\end{table}

TABLE~\ref{tab:ab_2} presents the ablation study to investigate the effect of each proposed module, such as SD, CAT, Synthetic EndoVis 2018 train dataset, and Synthetic pseudo-exemplars. Without the help of our synthetic framework, our CAT-SD approach improves the performance in regular, VS, and GR classes by 2.74\%, 2.92\%, and 0.34\% when using the real EndoVis 2018 dataset. However, due to the extreme scarcity of new class samples, the performance of Suction and CA stuck at 0. To alleviate the scarcity issue of new classes, we synthesize the EndoVis 2018 train dataset and find that the performance of Suction and CA is improved by 10.80\% and 9.31\%. Therefore, based on the Synthetic EndoVis 2018 train dataset, we further explore the effect of SD and CAT modules. The results reveal the significant impact of SD followed by CAT. Eventually, our Synthetic CAT-SD strikes a balance between allowing for the learning of new classes (plasticity) and preventing forgetting of old classes (rigidity).

\begin{figure}[!h]
\centering
\includegraphics[width=1\linewidth]{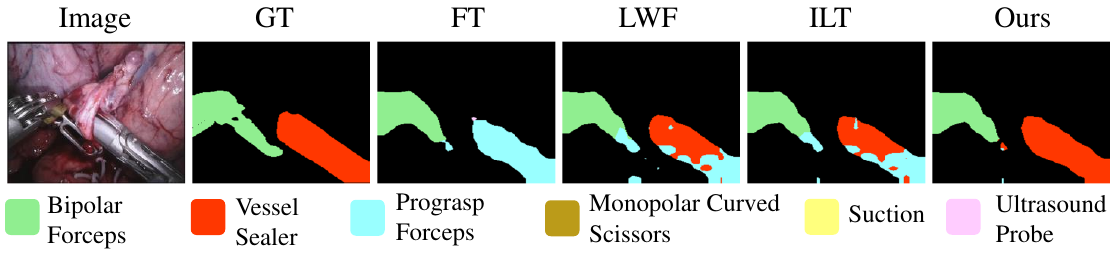}
\caption{Visualization of predicted segmentation on old class (Vessel Sealer) with our method over different baselines.}
\label{fig:predicted_old_class}
\end{figure}

Fig.~\ref{fig:predicted_old_class} shows the baseline CL methods (FT, LWF, and ILT) still present catastrophic forgetting for the \textit{old class (vessel sealer)}. These baseline CL methods have made error predictions on \textit{vessel sealer}, predicting the \textit{vessel sealer} (represented by the red color) as the \textit{prograsp forceps} (represented by the light blue). However, our approach can preserve the performance on the \textit{vessel sealer}.

\subsection{Robustness}
\change{The main objective of robustness verification is to assess the model's capacity to withstand uncertainties that may occur in real surgical situations. This is achieved by simulating the uncertainties and implementing changes to the dataset, following specific guidelines. Thus,} we apply perturbations in the images to investigate the robustness under input variations and observe the performance discrepancy. We design $3$ groups of perturbations with $5$ severity levels by following~\cite{hendrycks2019benchmarking}. More specifically, the perturbations are (i) Blur (defocus, glass, motion, zoom, Gaussian); (ii) Digital (contrast, elastic, pixel, jpeg); and (iii) Noise (Gaussian, Shot, Impulse, Speckle); (iv) Weather (snow, frost, fog, brightness); (v) Others (spatter, saturate, gamma), respectively. The severity levels are controlled by increasing the perturbation scale for each technique. For example, to add Gaussian noise with severity levels 1 to 5, the standard deviation of the Gaussian noise increase as [0.04, 0.06, 0.08, 0.09, 0.10], respectively. A model with better robustness will show higher IoU performance or preserve the performance of the clean data as the corruption severity level increases. The robustness performance in terms of fixing catastrophic forgetting of our approach is compared with methods of FT (without CL) and the closest baseline ILT~\cite{michieli2019ilt} and presented in the TABLE~\ref{tab:severity_level_specific}. Our method constantly preserves the higher performance across all the severity levels of the perturbations. \change{Our method incorporates diverse data augmentation techniques during the image synthesis process and leverages the robust feature learning of the CAT-SD approach to enhance its robustness and overall performance.}

\begin{table}[!h]
\centering
\caption{Robustness performance of the old instruments, vessel sealer (VS) and grasping retractor (GR) with 5 severity levels.}
\scalebox{0.75}{
\begin{tabular}{c|c|c|c|c|c|c|c|c|c|c} 
\hline
\multirow{2}{*}{Appraoch} & \multicolumn{2}{c|}{Severity-1}                      & \multicolumn{2}{c|}{Severity-2}                      & \multicolumn{2}{c|}{Severity-3}                      & \multicolumn{2}{c|}{Severity-4}                      & \multicolumn{2}{c}{Severity-5}        \\ 
\cline{2-11}
                          & VS                   & GR                   & VS                   & GR                   & VS                   & GR                   & VS                   & GR                   & VS                   & GR    \\ 
\hline
FT                        & 0.39                 & 3.02                 & 0.28                 & 2.38                 & 0.18                 & 2.11                 & 0.16                 & 1.56                 & 0.18                 & 1.03  \\
ILT                       & 11.69                & 4.03                 & 9.55                 & 3.42                 & 9.48                 & 3.21                 & 8.61                 & 2.35                 & 7.4                  & 1.51  \\ 
\hline
Ours                      & 15.27                & 4.28                 & 12.15                & 3.68                 & 11.66                & 3.32                 & 10.87                & 2.46                 & 8.87                 & 1.56  \\ 
\hline
\multicolumn{1}{c}{}      & \multicolumn{1}{c}{} & \multicolumn{1}{c}{} & \multicolumn{1}{c}{} & \multicolumn{1}{c}{} & \multicolumn{1}{c}{} & \multicolumn{1}{c}{} & \multicolumn{1}{c}{} & \multicolumn{1}{c}{} & \multicolumn{1}{c}{} &       \\
\multicolumn{1}{c}{}      & \multicolumn{1}{c}{} & \multicolumn{1}{c}{} & \multicolumn{1}{c}{} & \multicolumn{1}{c}{} & \multicolumn{1}{c}{} & \multicolumn{1}{c}{} & \multicolumn{1}{c}{} & \multicolumn{1}{c}{} & \multicolumn{1}{c}{} &       \\
\multicolumn{1}{c}{}      & \multicolumn{1}{c}{} & \multicolumn{1}{c}{} & \multicolumn{1}{c}{} & \multicolumn{1}{c}{} & \multicolumn{1}{c}{} & \multicolumn{1}{c}{} & \multicolumn{1}{c}{} & \multicolumn{1}{c}{} & \multicolumn{1}{c}{} &      
\end{tabular}
}
\label{tab:severity_level_specific}
\end{table}

\section{Discussion and Conclusions}
\label{sec5}
This work presents a privacy-preserving synthetic continual semantic segmentation approach to enable continual learning in robotic instrument segmentation. We developed a class-aware temperature-normalization-based multi-scale shifted distillation scheme (CAT-SD) that allows the model to preserve the learned information from the old non-overlapping classes and distill the knowledge from both short and long-range spatial relationships. We also integrate synthetic pseudo-exemplar to generate background tissue of the datasets to ensure privacy-preserving continual learning. In addition, we design a blending and harmonization module to blend synthetic surgical backgrounds with old instruments and real surgical backgrounds with new instruments with extensive augmentation. This facilitates the control of generating enough data to deal with expensive annotation, data scarcity, and patient privacy concerns. Our extensive experimental results reveal that (i) there is extreme catastrophic forgetting to fine-tune (FT) new instrument classes without CL; (ii) some traditional CL methods such as LWF-MC~\cite{rebuffi2017icarl}, ILT~\cite{michieli2019ilt} demonstrate the capacity to avoid the catastrophic forgetting but poorly learn for the regular and new instruments; (iii) our synthetic CAT-SD successfully learn to segment new instruments by preserving knowledge for the old instruments. To our knowledge, this is the first work in resolving continual learning issues of (i) catastrophic forgetting, (ii) learning dominance from overlapping classes, and (iii) privacy concerns in robotic surgery. Hence, to highlight the effectiveness of our method, we compare the performance with other SOTA continual learning techniques instead of other works with naive instrument segmentation tasks. Overall, the Synthetic CAT-SD approach results in a proper trade-off between rigidity and plasticity for continual semantic segmentation, eventually alleviating catastrophic forgetting, data shortage, data annotation, and privacy concerns. Our robustness test under input perturbations demonstrates that the proposed method can learn new knowledge and is robust under input variations.

\textbf{Future work:} When synthesizing images, we will achieve incremental domain adaptation by introducing (i) discrete domain shifts: the discrete domain parameters can be designed to generate different organs, for example, liver and kidneys as background images; (ii) continuous domain shifts: the domain parameters can be changed continuously to generate the images which gradually goes from clear to blood-filled, or smoke-filled.


\bibliographystyle{IEEEtran}
\bibliography{mybib}

\begin{thebibliography}{10}
\providecommand{\url}[1]{#1}
\csname url@samestyle\endcsname
\providecommand{\newblock}{\relax}
\providecommand{\bibinfo}[2]{#2}
\providecommand{\BIBentrySTDinterwordspacing}{\spaceskip=0pt\relax}
\providecommand{\BIBentryALTinterwordstretchfactor}{4}
\providecommand{\BIBentryALTinterwordspacing}{\spaceskip=\fontdimen2\font plus
\BIBentryALTinterwordstretchfactor\fontdimen3\font minus \fontdimen4\font\relax}
\providecommand{\BIBforeignlanguage}[2]{{%
\expandafter\ifx\csname l@#1\endcsname\relax
\typeout{** WARNING: IEEEtran.bst: No hyphenation pattern has been}%
\typeout{** loaded for the language `#1'. Using the pattern for}%
\typeout{** the default language instead.}%
\else
\language=\csname l@#1\endcsname
\fi
#2}}
\providecommand{\BIBdecl}{\relax}
\BIBdecl

\bibitem{gonzalez2020isinet}
C.~Gonz{\'a}lez, L.~Bravo-S{\'a}nchez, and P.~Arbelaez, ``Isinet: an instance-based approach for surgical instrument segmentation,'' in \emph{International Conference on Medical Image Computing and Computer-Assisted Intervention}.\hskip 1em plus 0.5em minus 0.4em\relax Springer, 2020, pp. 595--605.

\bibitem{mccloskey1989catastrophic}
M.~McCloskey and N.~J. Cohen, ``Catastrophic interference in connectionist networks: The sequential learning problem,'' in \emph{Psychology of learning and motivation}.\hskip 1em plus 0.5em minus 0.4em\relax Elsevier, 1989, vol.~24, pp. 109--165.

\bibitem{rebuffi2017icarl}
S.-A. Rebuffi, A.~Kolesnikov, G.~Sperl, and C.~H. Lampert, ``icarl: Incremental classifier and representation learning,'' in \emph{Proceedings of the IEEE conference on Computer Vision and Pattern Recognition}, 2017, pp. 2001--2010.

\bibitem{chaudhry2018riemannian}
A.~Chaudhry, P.~K. Dokania, T.~Ajanthan, and P.~H. Torr, ``Riemannian walk for incremental learning: Understanding forgetting and intransigence,'' in \emph{Proceedings of the European Conference on Computer Vision (ECCV)}, 2018, pp. 532--547.

\bibitem{li2017lwf}
Z.~Li and D.~Hoiem, ``Learning without forgetting,'' \emph{IEEE transactions on pattern analysis and machine intelligence}, vol.~40, no.~12, pp. 2935--2947, 2017.

\bibitem{douillard2020podnet}
A.~Douillard, M.~Cord, C.~Ollion, T.~Robert, and E.~Valle, ``Podnet: Pooled outputs distillation for small-tasks incremental learning,'' in \emph{European Conference on Computer Vision}.\hskip 1em plus 0.5em minus 0.4em\relax Springer, 2020, pp. 86--102.

\bibitem{douillard2021plop}
A.~Douillard, Y.~Chen, A.~Dapogny, and M.~Cord, ``Plop: Learning without forgetting for continual semantic segmentation,'' in \emph{Proceedings of the IEEE/CVF Conference on Computer Vision and Pattern Recognition}, 2021, pp. 4040--4050.

\bibitem{nikolenko2021synthetic}
S.~I. Nikolenko, \emph{Synthetic data for deep learning}.\hskip 1em plus 0.5em minus 0.4em\relax Springer, 2021, vol. 174.

\bibitem{liang2022advances}
W.~Liang, G.~A. Tadesse, D.~Ho, F.-F. Li, M.~Zaharia, C.~Zhang, and J.~Zou, ``Advances, challenges and opportunities in creating data for trustworthy ai,'' \emph{Nature Machine Intelligence}, pp. 1--9, 2022.

\bibitem{van2020ethical}
R.~Van~Noorden, ``The ethical questions that haunt facial-recognition research,'' \emph{Nature}, vol. 587, no. 7834, pp. 354--359, 2020.

\bibitem{buolamwini2018gender}
J.~Buolamwini and T.~Gebru, ``Gender shades: Intersectional accuracy disparities in commercial gender classification,'' in \emph{Conference on fairness, accountability and transparency}.\hskip 1em plus 0.5em minus 0.4em\relax PMLR, 2018, pp. 77--91.

\bibitem{kortylewski2019analyzing}
A.~Kortylewski, B.~Egger, A.~Schneider, T.~Gerig, A.~Morel-Forster, and T.~Vetter, ``Analyzing and reducing the damage of dataset bias to face recognition with synthetic data,'' in \emph{Proceedings of the IEEE/CVF Conference on Computer Vision and Pattern Recognition Workshops}, 2019, pp. 0--0.

\bibitem{wang2022rethinking}
A.~Wang, M.~Islam, M.~Xu, and H.~Ren, ``Rethinking surgical instrument segmentation: A background image can be all you need,'' in \emph{International Conference on Medical Image Computing and Computer-Assisted Intervention}.\hskip 1em plus 0.5em minus 0.4em\relax Springer, 2022, pp. 355--364.

\bibitem{garcia2021image}
L.~C. Garcia-Peraza-Herrera, L.~Fidon, C.~D’Ettorre, D.~Stoyanov, T.~Vercauteren, and S.~Ourselin, ``Image compositing for segmentation of surgical tools without manual annotations,'' \emph{IEEE transactions on medical imaging}, vol.~40, no.~5, pp. 1450--1460, 2021.

\bibitem{mai2021supervised}
Z.~Mai, R.~Li, H.~Kim, and S.~Sanner, ``Supervised contrastive replay: Revisiting the nearest class mean classifier in online class-incremental continual learning,'' in \emph{Proceedings of the IEEE/CVF Conference on Computer Vision and Pattern Recognition}, 2021, pp. 3589--3599.

\bibitem{shin2017continual}
H.~Shin, J.~K. Lee, J.~Kim, and J.~Kim, ``Continual learning with deep generative replay,'' \emph{Advances in neural information processing systems}, vol.~30, 2017.

\bibitem{liu2020generative}
X.~Liu, C.~Wu, M.~Menta, L.~Herranz, B.~Raducanu, A.~D. Bagdanov, S.~Jui, and J.~v. de~Weijer, ``Generative feature replay for class-incremental learning,'' in \emph{Proceedings of the IEEE/CVF Conference on Computer Vision and Pattern Recognition Workshops}, 2020, pp. 226--227.

\bibitem{wu2018incremental}
Y.~Wu, Y.~Chen, L.~Wang, Y.~Ye, Z.~Liu, Y.~Guo, Z.~Zhang, and Y.~Fu, ``Incremental classifier learning with generative adversarial networks,'' \emph{arXiv preprint arXiv:1802.00853}, 2018.

\bibitem{maracani2021recall}
A.~Maracani, U.~Michieli, M.~Toldo, and P.~Zanuttigh, ``Recall: Replay-based continual learning in semantic segmentation,'' in \emph{Proceedings of the IEEE/CVF International Conference on Computer Vision}, 2021, pp. 7026--7035.

\bibitem{oquab2014learning}
M.~Oquab, L.~Bottou, I.~Laptev, and J.~Sivic, ``Learning and transferring mid-level image representations using convolutional neural networks,'' in \emph{Proceedings of the IEEE conference on computer vision and pattern recognition}, 2014, pp. 1717--1724.

\bibitem{sarwar2019incremental}
S.~S. Sarwar, A.~Ankit, and K.~Roy, ``Incremental learning in deep convolutional neural networks using partial network sharing,'' \emph{IEEE Access}, vol.~8, pp. 4615--4628, 2019.

\bibitem{xiao2014error}
T.~Xiao, J.~Zhang, K.~Yang, Y.~Peng, and Z.~Zhang, ``Error-driven incremental learning in deep convolutional neural network for large-scale image classification,'' in \emph{Proceedings of the 22nd ACM international conference on Multimedia}, 2014, pp. 177--186.

\bibitem{hinton2015distilling}
G.~Hinton, O.~Vinyals, J.~Dean \emph{et~al.}, ``Distilling the knowledge in a neural network,'' \emph{arXiv preprint arXiv:1503.02531}, vol.~2, no.~7, 2015.

\bibitem{bucilu2006model}
C.~Buciluǎ, R.~Caruana, and A.~Niculescu-Mizil, ``Model compression,'' in \emph{Proceedings of the 12th ACM SIGKDD international conference on Knowledge discovery and data mining}, 2006, pp. 535--541.

\bibitem{michieli2019ilt}
U.~Michieli and P.~Zanuttigh, ``Incremental learning techniques for semantic segmentation,'' in \emph{Proceedings of the IEEE/CVF International Conference on Computer Vision Workshops}, 2019, pp. 0--0.

\bibitem{mi2020generalized}
F.~Mi, L.~Kong, T.~Lin, K.~Yu, and B.~Faltings, ``Generalized class incremental learning,'' in \emph{Proceedings of the IEEE/CVF Conference on Computer Vision and Pattern Recognition Workshops}, 2020, pp. 240--241.

\bibitem{shin2018medical}
H.-C. Shin, N.~A. Tenenholtz, J.~K. Rogers, C.~G. Schwarz, M.~L. Senjem, J.~L. Gunter, K.~P. Andriole, and M.~Michalski, ``Medical image synthesis for data augmentation and anonymization using generative adversarial networks,'' in \emph{International workshop on simulation and synthesis in medical imaging}.\hskip 1em plus 0.5em minus 0.4em\relax Springer, 2018, pp. 1--11.

\bibitem{hamghalam2020high}
M.~Hamghalam, B.~Lei, and T.~Wang, ``High tissue contrast mri synthesis using multi-stage attention-gan for segmentation,'' in \emph{Proceedings of the AAAI conference on artificial intelligence}, vol.~34, no.~04, 2020, pp. 4067--4074.

\bibitem{lee2019davincigan}
K.~Lee, M.-K. Choi, and H.~Jung, ``Davincigan: Unpaired surgical instrument translation for data augmentation,'' in \emph{International Conference on Medical Imaging with Deep Learning}.\hskip 1em plus 0.5em minus 0.4em\relax PMLR, 2019, pp. 326--336.

\bibitem{pfeiffer2019generating}
M.~Pfeiffer, I.~Funke, M.~R. Robu, S.~Bodenstedt, L.~Strenger, S.~Engelhardt, T.~Ro{\ss}, M.~J. Clarkson, K.~Gurusamy, B.~R. Davidson \emph{et~al.}, ``Generating large labeled data sets for laparoscopic image processing tasks using unpaired image-to-image translation,'' in \emph{Medical Image Computing and Computer Assisted Intervention--MICCAI 2019: 22nd International Conference, Shenzhen, China, October 13--17, 2019, Proceedings, Part V 22}.\hskip 1em plus 0.5em minus 0.4em\relax Springer, 2019, pp. 119--127.

\bibitem{rivoir2021long}
D.~Rivoir, M.~Pfeiffer, R.~Docea, F.~Kolbinger, C.~Riediger, J.~Weitz, and S.~Speidel, ``Long-term temporally consistent unpaired video translation from simulated surgical 3d data,'' in \emph{Proceedings of the IEEE/CVF International Conference on Computer Vision}, 2021, pp. 3343--3353.

\bibitem{yoon2022surgical}
J.~Yoon, S.~Hong, S.~Hong, J.~Lee, S.~Shin, B.~Park, N.~Sung, H.~Yu, S.~Kim, S.~Park \emph{et~al.}, ``Surgical scene segmentation using semantic image synthesis with a virtual surgery environment,'' in \emph{Medical Image Computing and Computer Assisted Intervention--MICCAI 2022: 25th International Conference, Singapore, September 18--22, 2022, Proceedings, Part VII}.\hskip 1em plus 0.5em minus 0.4em\relax Springer, 2022, pp. 551--561.

\bibitem{colleoni2020synthetic}
E.~Colleoni, P.~Edwards, and D.~Stoyanov, ``Synthetic and real inputs for tool segmentation in robotic surgery,'' in \emph{Medical Image Computing and Computer Assisted Intervention--MICCAI 2020: 23rd International Conference, Lima, Peru, October 4--8, 2020, Proceedings, Part III 23}.\hskip 1em plus 0.5em minus 0.4em\relax Springer, 2020, pp. 700--710.

\bibitem{ding2022carts}
H.~Ding, J.~Zhang, P.~Kazanzides, J.~Y. Wu, and M.~Unberath, ``Carts: Causality-driven robot tool segmentation from vision and kinematics data,'' in \emph{Medical Image Computing and Computer Assisted Intervention--MICCAI 2022: 25th International Conference, Singapore, September 18--22, 2022, Proceedings, Part VII}.\hskip 1em plus 0.5em minus 0.4em\relax Springer, 2022, pp. 387--398.

\bibitem{allan20192017}
M.~Allan, A.~Shvets, T.~Kurmann, Z.~Zhang, R.~Duggal, Y.-H. Su, N.~Rieke, I.~Laina, N.~Kalavakonda, S.~Bodenstedt \emph{et~al.}, ``2017 robotic instrument segmentation challenge,'' \emph{arXiv preprint arXiv:1902.06426}, 2019.

\bibitem{allan20202018}
M.~Allan, S.~Kondo, S.~Bodenstedt, S.~Leger, R.~Kadkhodamohammadi, I.~Luengo, F.~Fuentes, E.~Flouty, A.~Mohammed, M.~Pedersen \emph{et~al.}, ``2018 robotic scene segmentation challenge,'' \emph{arXiv preprint arXiv:2001.11190}, 2020.

\bibitem{dhar2019learning}
P.~Dhar, R.~V. Singh, K.-C. Peng, Z.~Wu, and R.~Chellappa, ``Learning without memorizing,'' in \emph{Proceedings of the IEEE/CVF conference on computer vision and pattern recognition}, 2019, pp. 5138--5146.

\bibitem{zhou2019m2kd}
P.~Zhou, L.~Mai, J.~Zhang, N.~Xu, Z.~Wu, and L.~S. Davis, ``M2kd: Multi-model and multi-level knowledge distillation for incremental learning,'' \emph{arXiv preprint arXiv:1904.01769}, 2019.

\bibitem{jaynes1957information}
E.~T. Jaynes, ``Information theory and statistical mechanics,'' \emph{Physical review}, vol. 106, no.~4, p. 620, 1957.

\bibitem{guo2017calibration}
C.~Guo, G.~Pleiss, Y.~Sun, and K.~Q. Weinberger, ``On calibration of modern neural networks,'' in \emph{International Conference on Machine Learning}.\hskip 1em plus 0.5em minus 0.4em\relax PMLR, 2017, pp. 1321--1330.

\bibitem{liu2021swin}
Z.~Liu, Y.~Lin, Y.~Cao, H.~Hu, Y.~Wei, Z.~Zhang, S.~Lin, and B.~Guo, ``Swin transformer: Hierarchical vision transformer using shifted windows,'' in \emph{Proceedings of the IEEE/CVF International Conference on Computer Vision}, 2021, pp. 10\,012--10\,022.

\bibitem{song2022spot}
J.~Song, Y.~Chen, J.~Ye, and M.~Song, ``Spot-adaptive knowledge distillation,'' \emph{IEEE Transactions on Image Processing}, vol.~31, pp. 3359--3370, 2022.

\bibitem{hendrycks2019benchmarking}
D.~Hendrycks and T.~Dietterich, ``Benchmarking neural network robustness to common corruptions and perturbations,'' \emph{arXiv preprint arXiv:1903.12261}, 2019.

\bibitem{sofiiuk2021foreground}
K.~Sofiiuk, P.~Popenova, and A.~Konushin, ``Foreground-aware semantic representations for image harmonization,'' in \emph{Proceedings of the IEEE/CVF Winter Conference on Applications of Computer Vision}, 2021, pp. 1620--1629.

\bibitem{masana2020class}
M.~Masana, X.~Liu, B.~Twardowski, M.~Menta, A.~D. Bagdanov, and J.~van~de Weijer, ``Class-incremental learning: survey and performance evaluation on image classification,'' \emph{arXiv preprint arXiv:2010.15277}, 2020.

\bibitem{allan20183}
M.~Allan, S.~Ourselin, D.~J. Hawkes, J.~D. Kelly, and D.~Stoyanov, ``3-d pose estimation of articulated instruments in robotic minimally invasive surgery,'' \emph{IEEE transactions on medical imaging}, vol.~37, no.~5, pp. 1204--1213, 2018.

\bibitem{sauer2022stylegan}
A.~Sauer, K.~Schwarz, and A.~Geiger, ``Stylegan-xl: Scaling stylegan to large diverse datasets,'' in \emph{Special Interest Group on Computer Graphics and Interactive Techniques Conference Proceedings}, 2022, pp. 1--10.

\bibitem{chen2017rethinking}
L.-C. Chen, G.~Papandreou, F.~Schroff, and H.~Adam, ``Rethinking atrous convolution for semantic image segmentation,'' \emph{arXiv preprint arXiv:1706.05587}, 2017.

\bibitem{he2016deep}
K.~He, X.~Zhang, S.~Ren, and J.~Sun, ``Deep residual learning for image recognition,'' in \emph{Proceedings of the IEEE conference on computer vision and pattern recognition}, 2016, pp. 770--778.

\bibitem{zenke2017si}
F.~Zenke, B.~Poole, and S.~Ganguli, ``Continual learning through synaptic intelligence,'' in \emph{International Conference on Machine Learning}.\hskip 1em plus 0.5em minus 0.4em\relax PMLR, 2017, pp. 3987--3995.

\bibitem{kamil2019literature}
A.~Kamil and T.~Shaikh, ``Literature review of generative models for image-to-image translation problems,'' in \emph{2019 International Conference on Computational Intelligence and Knowledge Economy (ICCIKE)}.\hskip 1em plus 0.5em minus 0.4em\relax IEEE, 2019, pp. 340--345.

\bibitem{ho2020denoising}
J.~Ho, A.~Jain, and P.~Abbeel, ``Denoising diffusion probabilistic models,'' \emph{Advances in neural information processing systems}, vol.~33, pp. 6840--6851, 2020.

\end{thebibliography}

\end{document}